\documentclass[journal,doublecolumn]{IEEEtran}
\usepackage{graphicx}
\usepackage{subcaption}
\usepackage{amsmath}
\usepackage{amsfonts}
\usepackage{bbm}
\usepackage{booktabs}
\usepackage{hyperref}
\usepackage{bm}
\usepackage{xcolor}
\usepackage{lineno}

\newcommand{\ical}[1]{{\mbox{\usefont{OT1}{pzc}{m}{it}{#1}}}}
\newcommand{\trans}[0]{^{\sf T}}
\newcommand{\m}[1]{{\mbox{{\fontencoding{T1}\sffamily\slshape{#1\/}}}}}
\renewcommand{\d}[1]{{\mbox{\boldmath$#1$}}}

\begin{document}

\title{\textcolor{black}{ Confident Naturalness Explanation (CNE): A Framework to Explain and Assess Patterns Forming Naturalness }}

\author{Ahmed~Emam~\IEEEmembership{,}
        Mohamed~Farag~\IEEEmembership{,}
        and~Ribana~Roscher,~\IEEEmembership{Member,~IEEE}
\thanks{A. Emam and M. Farag are with the Institute
of Geodesy and Geoinformation, University of Bonn, 53113 Bonn, Germany. (aemam,~mibrahi2)@uni-bonn.de}
\thanks{R. Roscher is with the Institute of Bio- and Geosciences, Forschungszentrum Jülich, 52428 Jülich, Germany, and the Institute
of Geodesy and Geoinformation, University of Bonn, 53113 Bonn, Germany.~(r.roscher@fz-juelich.de)}
\thanks{Manuscript received; revised }}

\markboth{Journal of \LaTeX\ Class Files,~Vol.~, No.~, February~2023}%
{Emam \MakeLowercase{\textit{et al.}}: \textcolor{black}{Confident Naturalness Explanation (CNE)}: A Framework to Explain and Assess Patterns Forming Naturalness}

\maketitle



\IEEEpeerreviewmaketitle

\begin{abstract}
    
\IEEEPARstart{P}{rotected }\textcolor{black}{ natural areas characterized by minimal modern human footprint are often challenging to assess. Machine learning models, particularly explainable methods, offer promise in understanding and mapping the naturalness of these environments through the analysis of satellite imagery. However, current approaches encounter challenges in delivering valid and objective explanations and quantifying the contribution of specific patterns to naturalness. These challenges persist due to the reliance on hand-crafted weights assigned to contributing patterns, which can introduce subjectivity and limit the model's ability to capture relationships within the data.}
 \textcolor{black}{We propose the Confident Naturalness Explanation (CNE) framework to address these issues, integrating explainable machine learning and uncertainty quantification. This framework introduces a new quantitative metric to describe the confident contribution of patterns to the concept of naturalness. Additionally, it generates segmentation masks that depict the uncertainty levels in each pixel, highlighting areas where the model lacks knowledge.}
 \textcolor{black}{To showcase the framework's effectiveness, we apply it to a study site in Fennoscandia, utilizing two open-source satellite datasets. In our proposed metric scale, moors and heathlands register high values of 1 and 0.81, respectively, indicating pronounced naturalness. In contrast, water bodies score lower on the scale, with a metric value of 0.18, placing them at the lower end.}
\end{abstract}
\begin{figure*}[t]
    \centering
    \includegraphics[trim = 0cm 0cm 0cm 1cm, width=\textwidth]{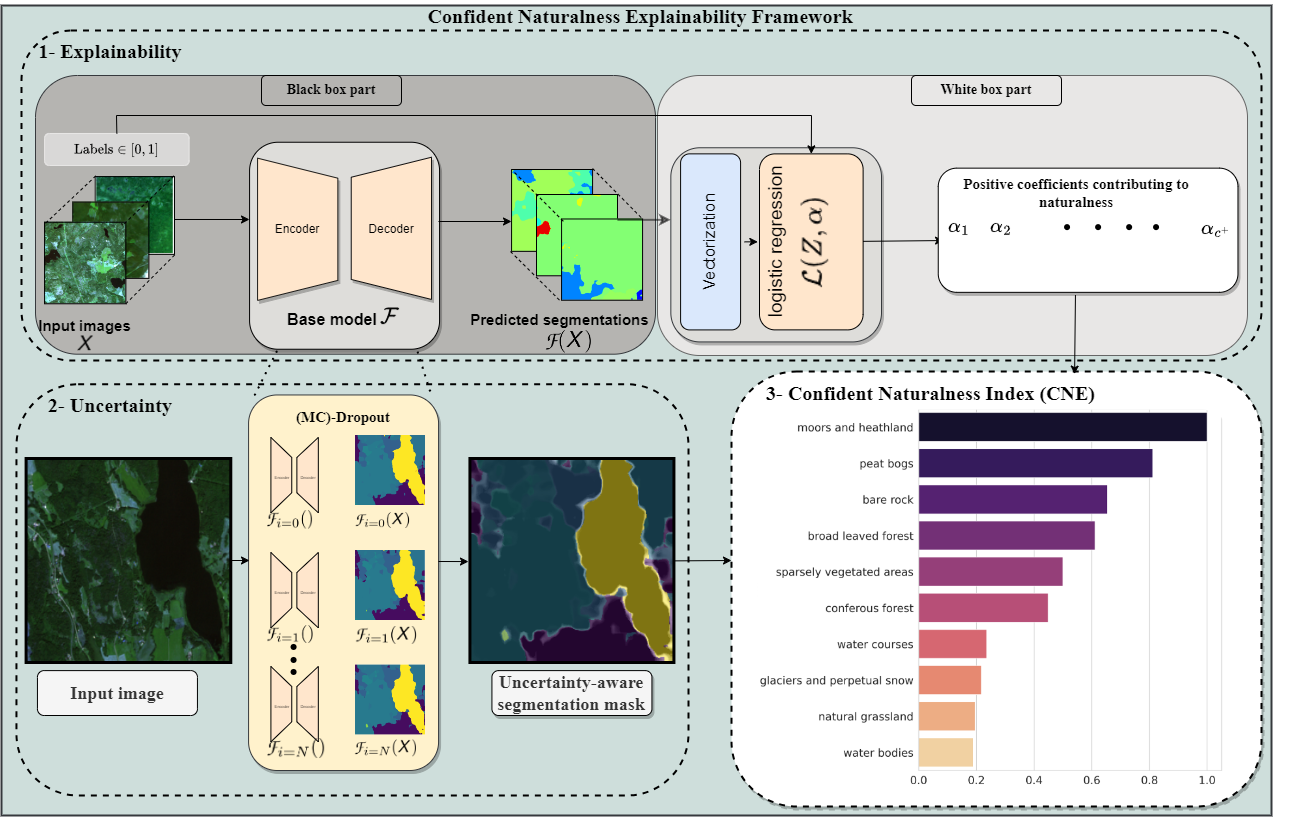}
    \caption{\textbf{An Illustration for (CNE) framework.} In the explainability part, the input images are fed to the segmentation model, resulting in predicted segmentation masks; they are fed to logistic regression with ground truth labels of the input images to vectorize patterns and classify the input into naturalness and anthropogenic areas. \textcolor{black}{After training the logistic regression, we use only the positive coefficients to calculate the CNE metric}. \textcolor{black}{In the uncertainty, the MC-Dropout resembles multiple sampled models used to quantify the uncertainty of each pattern in the input image. In the lower right corner, the knowledge gained from parts 1 and 2 is combined to calculate the CNE metric in part 3} and assign a quantifiable metric value to each pattern, reflecting its confident contribution to the concept of naturalness. \textcolor{black}{ The uncertainty part is shown in detail in \hyperref[fig2:output]{Fig. 2}}}
    \label{fig:framework}
\end{figure*}
\section{Introduction}
Protected natural areas are regions of the Earth that have remained largely untouched by significant human intervention, such as urbanization, agriculture, and other human activities \cite{sze_reduced_2022}. These areas are characterized by their preserved and authentic state, and they boast high levels of biodiversity and provide numerous ecological benefits. They also offer unique opportunities to study natural ecosystem processes, including water and pollination cycles, in their unaltered form.
To maintain the authenticity of these areas, it is crucial to conduct careful and comprehensive mapping and monitoring efforts. One obstacle, however, is the vague definition of the land cover class, which prevents a comprehensive mapping. Nevertheless, these efforts help reveal the complex geo-ecological patterns essential for preserving these regions' naturalness. As a result, the monitoring and understanding of natural areas have gained significant attention in both remote sensing and environmental research fields recently \cite{mittermeier_wilderness_2003,smith_strategic_2021}. 
Satellite imagery emerges as an effective technique for consistently observing vast protected natural expanses, which can be challenging for humans to access. This technology \textcolor{black}{enables} efficient and cost-effective data collection while minimizing disturbances to delicate ecosystems.
\textcolor{black}{By using} Machine Learning (ML) models, \textcolor{black}{specifically} Convolutional Neural Networks (CNNs), \textcolor{black}{it becomes possible to accurately classify} natural regions by analyzing satellite imagery datasets. 


\textcolor{black}{In previous studies on naturalness analysis, such as \cite{sanderson_human_2002} and \cite{ekim_naturalness_2021}, the quantification of naturalness has been done by defining a feature-engineered modern human footprint index within a predefined range of 0 to 10. These studies use proxies like railways, electric power structures, and population density to assess the impact of human activity. Our approach distinguishes itself by specifically focusing on naturalness as the primary concept of interest and assessing and explaining the underlying patterns that contribute to it. With a similar goal, though a different approach,}  \textcolor{black}{Explanatory frameworks designed by \cite{stomberg_exploring_2022, emam_leveraging_2024} generate attribution maps that effectively highlight patterns indicative of protected natural areas in satellite imagery}
However, despite the effectiveness of these methods in identifying \textcolor{black}{natural regions using specific indices} or distinct patterns characterizing natural regions, \textcolor{black}{they} struggle to provide a quantitative metric that accurately reflects the patterns' importance\textcolor{black}{, while also considering their} certainty.

To address these limitations, \textcolor{black}{we propose a framework that extracts patterns that contribute to the concept of naturalness in satellite imagery and assesses the importance and certainty of these patterns with a novel metric called \textcolor{black}{Confident Naturalness Explanation (CNE)}. This metric combines existing tools for explainability and uncertainty quantification that} enables prioritization and sorting of the contributing patterns based on their quality. \textcolor{black}{Our main contributions are:
\begin{enumerate}
    \item \textcolor{black}{We demonstrate our CNE framework using a study site in Fennoscandia. We extract patterns associated with the concept of naturalness and protected areas from the AnthroProtect dataset \cite{stomberg_exploring_2022} by using domain knowledge from the comprehensive, well-known, and well-understood CORINE data set \cite{european_environment_agency_corine_2019}.} 
    \item \textcolor{black}{We generate u}ncertainty-aware segmentation masks to highlight the pixels where the model exhibits the lowest certainty.
    \item \textcolor{black}{We compute CNE,} merging both explainability and uncertainty to show the significance of each pattern to the concept of naturalness.
\end{enumerate}
}
\textcolor{black}{Here, we use the term ``pattern" for a land cover class. However, our framework is versatile in this regard, and the term can be used more broadly, for example, for a temporal or spatial regularity in a data set, as already proposed in \cite{sanderson_human_2002}.}


\textcolor{black}{\section{Confident Naturalness Explanation Framework}}
In the following, the CNE framework (\hyperref[fig:framework]{Fig. 1}) is presented in the context of our study site in Fennoscandia. 
\subsection{Study Site and Data Sets}
The goal of the CNE framework is to extract patterns from a data set and explain them by utilizing domain knowledge from another source of information. Here, we intend to find patterns associated with the concept of naturalness in the coarse AnthroProtect dataset \cite{stomberg_exploring_2022} with the help of the more nuanced, well-understood CORINE dataset \cite{european_environment_agency_corine_2019}.
The AnthroProtect dataset comprises 24,000 multispectral Sentinel-2 images of the $256 \times 256$ pixels in the Fennoscandia region. In our study, we focus on the red, green, and blue bands. The reference images are either classified as protected (WDPA categories: ``strict nature reserve" (Ia), ``wilderness" (Ib), ``national park" (II)) or anthropogenic areas. The protected areas are used as a proxy for naturalness; by doing so, we aim to establish a foundation for understanding and explaining naturalness, starting with these well-protected and minimally influenced environments. 
Anthropogenic areas in the AnthroProtect dataset align closely with ``artificial surfaces" and ``agricultural areas" in the CORINE land cover classes (patterns), which means that both data sets are consistent. CORINE dataset's unique composition captures the broad spectrum of naturalness in Fennoscandia. 
The original (1 $\times H \times W$)-dimensional label masks are converted to a ($C \times H \times W$) one-hot-encoded mask, where $H \times W$ is the size of the image, and $C$ is set to 44, creating a channel for each class from the original label. This results in 1s and 0s in these channels, with the rest as 0s.

\textcolor{black}{\subsection{Components of the Framework}}
The CNE framework shown in \hyperref[fig:framework]{Fig. 1} consists of \textcolor{black}{three} parts, detailed below. \textcolor{black}{In the first part, the explainability part, we train a black-box semantic segmentation model with the CORINE land cover segmentation masks. We} assign an importance value to pre-defined patterns \textcolor{black}{by learning a white-box logistic regression model that uses as input the segmentation model output and the AnthroProtect labels. Both together build a grey-box approach that is inspired by \cite{bennetot_greybox_2022}.}   In the \textcolor{black}{second} part, the Monte Carlo (MC)-Dropout technique is used to quantify the uncertainty in predicting the patterns contributing to naturalness. The gained knowledge is integrated to create the CNE metric \textcolor{black}{in the third part}, which assigns confident importance to the patterns forming naturalness in Fennoscandia.

\begin{figure*}[ht]
    \centering
    \includegraphics[trim = 0cm 0.5cm 0cm 1cm, width=\textwidth, height=7cm]{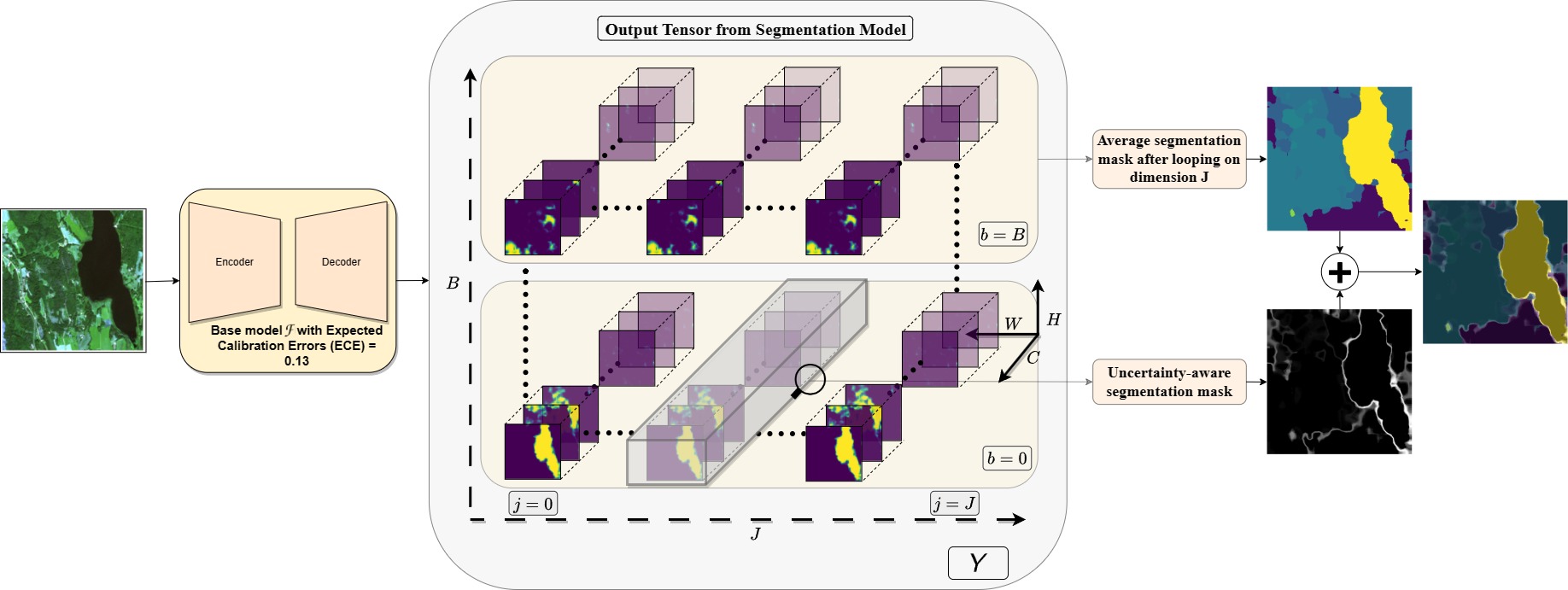}
    \caption{\textbf{Illustrative diagram for the tensor generated by the segmentation model.} The input images are passed to the model, and the middlebox includes the tensor $\m Y$ where $b$ is the image index at a single batch, and $j$ is the number of sampled models. \textcolor{black}{On the right, we have two outputs. The upper image \textcolor{black}{shows the output after taking the average over dimension $J$ and assigning each pixel to the class with the highest probability}, and at the bottom, \textcolor{black}{an uncertainty aware segmentation mask is generated by getting the standard deviation across MC runs where high-intensity pixels -white- represent high uncertainty and vice-versa}.}}
    \label{fig2:output}
\end{figure*}

\textcolor{black}{\textbf{(1) Explainability.}}
The first part of the grey box approach is a \textcolor{black}{black-box} semantic segmentation model $\ical F$. In this work, we utilize the DeepLabV3 based on ResNet-50 backbone \cite{Chen2017}. It efficiently captures multi-scale contextual information, enabling precise object boundary delineation. DeepLabV3 uses varying dilation rates to comprehend intricate image details and context through its spatial pyramid pooling module, resulting in highly accurate and fine-grained segmentation outcomes \cite{chen_rethinking_2017}. The architecture has a dropout layer localized after the last fully connected layers and before the output layer. The layer in our architecture serves a dual purpose: It contributes to regularization during training and plays a crucial role in uncertainty quantification. By employing MC-Dropout \cite{Gal2015}, we use the dropout layer to evaluate prediction uncertainty, offering insights into the model's confidence.
The first prime component of the grey box resembles as
\begin{equation}
\label{segmentation}
\ical F (\m X) = \m Y.
\end{equation}
In our case,  the output $\m Y ^{B \times J \times C \times H \times W}$ is a $5$-dimensionsal tensor, with $B$ being the number of images, $H$ and $W$ being the width and height of each image, $C$ being the number of classes, and J the number of MC-Dropout runs as shown in \hyperref[fig2:output]{Fig. 2}. For simplicity, we omit the indices $b$ for the image $j$ for the MC dropout run for further consideration unless we consider multiple images and runs. Each predicted segmentation mask $\m Y$ comprises $C$ binary masks, one for each class, indicating which pixels are classified as class $c$. The segmentation mask is transformed into a $C\times1$-dimensional vector of patterns $\d z$, where each element in the vector represents the abundance of a specific pattern in the predicted segmentation maps. The segmentation mask vectorization is described as follows:
\begin{equation}
\label{vectorization}
    \d z_{c} = \sum_{h,w} \m Y_{w,h,c}\,, 
\end{equation}
which means that all pixels that are predicted as class $c$ in each binary segmentation mask are summed.


\textcolor{black}{In the second part of the grey box approach}, we employ logistic regression, known for its high interpretability and alignment with algorithmic transparency criteria. It optimizes coefficients to classify predicted segmentation masks as ``naturalness'' or ``anthropogenic''. \textcolor{black}{It's worth noting that the grey box approach may sacrifice some spatial information, but this trade-off serves our primary goal of achieving explainable results.} 

For this, a logistic regression classification model $\ical L (\d z;\bm{\alpha})$ is used 
\begin{equation} \label{log_reg}
\ical L (\d z; \bm{\alpha}) = \frac{1}{1+e^{-\alpha\trans\d z}}\,,
\end{equation}
with $\bm{\alpha}$ being a $C$-dimensional vector where each coefficient is assigned to one class $c$.

\textcolor{black}{\textbf{(2) Uncertainty Quantification:}}
Improving models by providing additional information about outputs to increase confidence \textcolor{black}{plays an important} role in uncertainty estimation. Models approximate the real world, creating two sources of uncertainty: \textit{Epistemic} uncertainty (model uncertainty) and \textit{Aleatoric} uncertainty (data uncertainty) \cite{Huellermeier2019}. Model uncertainty represents the lack of knowledge about the best model, while data uncertainty relates to the inherent stochastic component in the data-generating process. \textcolor{black}{Predictive uncertainty is the combination of both.}
\textcolor{black}{A well-known method that provides aleatoric and epistemic} uncertainty \textcolor{black}{estimates is} Bayesian Neural Networks (BNN) \cite{Wang2020}. \textcolor{black}{Also} softmax outputs can be used to assess uncertainty in a fast way, but \textcolor{black}{they provide} only the aleatoric part. Set-valued predictive techniques like Conformal Prediction (CP) \cite{Vovk2005, Papadopoulos2002, Lei2014} are used for estimating the overall predictive uncertainty.
In \textcolor{black}{deep learning} (DL), the common framework is to obtain a point estimate of the model's weights. However, there is a growing need to estimate the model's knowledge, or epistemic uncertainty, which has led to the development of methods for addressing this issue. BNNs are used to generate weight distributions that capture the model's uncertainty. Nevertheless, due to the complexity of obtaining such distributions, approximate techniques such as MC-Dropout and ensemble learning \cite{Hoffmann2021} are widely adopted.

Dropout is a regularization approach utilized in DL to reduce overfitting \cite{he_deep_2015}. During training at each forward pass $j$, dropout randomly deactivates some of the neurons controlled by a hyperparameter $p_{\text{drop}}$, which represents the fraction of neurons to switch off to force the DL model to have distributed representations. During inference, the dropout layer is switched off to avoid getting stochastic outputs.

MC-Dropout exploits the previous idea to estimate the epistemic uncertainty by allowing the model to get a stochastic output during multiple forward runs $j=1,...,J$ at the inference phase. 
The mean prediction of each pattern $c$ for different model samples $\ical F_j$ is $\m A_{c}$ for a single image $\m X$ and obtained as
\begin{equation}
    \m A_{c}= \frac{1}{J}\sum_{j=1}^{J}\m Y_{c,j}.
\end{equation}
Furthermore, the standard deviations $\m S$ can be analyzed to check the change in pixel-level values generated per class:
\begin{equation}
\m S_{c} = \sqrt{\frac{1}{J}\sum_{j=1}^{J} (\m Y_{c,j} - \m A_{c}})^{2}.
\end{equation}

\begin{figure*}[htp]
    \centering
    \includegraphics[trim = 0cm 0.55cm 0cm 0.8 cm,width=\textwidth]{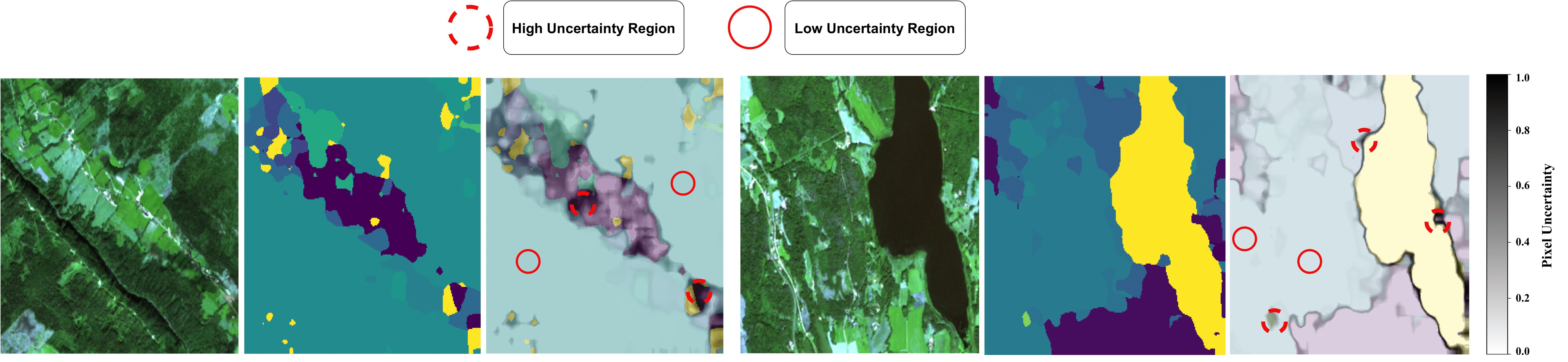}
    \caption{\textbf{Qualitative results.} \textcolor{black}{Demonstrating RGB Sentinel 2 images, predicted segmentation masks, and uncertainty-aware segmentation maps for two examples. The greyscale bar indicates pixel uncertainty in the segmentation maps. Each color in the segmentation masks represents a different pattern}}
    \label{fig:results2}
\end{figure*}

{\textbf{(3) \textcolor{black}{Confident} Naturalness Explanation Metric (CNE):}}\label{sec2.3}
We combine the knowledge gained in both the explainability and the uncertainty quantification parts and 
propose the CNE metric, which assesses the quality of explanations of the patterns contributing to the concept of naturalness in Fennoscandia.
The metric is bounded between 0 and 1, where the pattern that has a value of 1 contributes significantly to the concept of naturalness with high certainty, and a pattern that has a value of zero will either have a significantly low contribution to the concept of naturalness or high uncertainty.

The metric is calculated as follows:
\begin{equation}
\label{CNE metric}
\text{CNE}_{c} = \frac{\bm{\alpha}_{c^{+}}}{u_c}\,,
\end{equation}
with
\begin{equation*}
\label{alpha+}
    \bm{\alpha_{c^{+}}} = \max(\bm{\alpha_{c}, 0)},
\end{equation*}
\begin{equation*}
    u_c = \sum\limits_{h, w} \m S_c\,.
\end{equation*}
The term $\alpha_{c^{+}}$ represents the modified trained logistic regressor's coefficient at which negative values are set to zero. The \(\max\) function ensures that only positive coefficients are retained to include patterns that positively contribute to the naturalness concepts \textcolor{black}{to ensure valid and objective explanations of naturalness}. At the same time, $u$ is the summation of the patterns' pixels across the spatial dimensions after taking the standard deviation over the MC-dropout dimension $J$.


\section{Experiments, results, and discussions}

\subsection{Experimental setup}
In DeepLabV3 \cite{chen_rethinking_2017}, a dropout layer is placed after the last fully connected layers with $p_\text{drop}$ = 0.1. The model achieved 91.2\% IOU on the training set and 80.3\% on the test set after 100 epochs using 80\% of the AnthroProtect dataset.


\subsection{Results Interpretation}

\begin{table}
\centering
\begin{tabular}{ccc}
\toprule
\textcolor{black}{Pattern} & Metric & Distribution\%\\ 
\midrule
moors and heathland & 1.00 & \textcolor{black}{13.2} \\
peat bogs & 0.81 & \textcolor{black}{6.1} \\
bare rock & 0.65 & \textcolor{black}{6.8} \\
broad-leaved forest & 0.61 & \textcolor{black}{13.4} \\
sparsely vegetated areas & 0.49 & \textcolor{black}{24.1} \\
coniferous forest & 0.44 & \textcolor{black}{18.2} \\
watercourses & 0.23 & \textcolor{black}{0.2} \\
glaciers and perpetual snow & 0.21 & \textcolor{black}{1.1} \\
natural grassland & 0.19 & \textcolor{black}{0.05} \\
water bodies & 0.18 & \textcolor{black}{5.2} \\
\bottomrule
\end{tabular}
\caption{\textbf{CNE metric values}. Each pattern has a value representing its confident contribution to the concept of naturalness. \textcolor{black}{We exclude non-designating patterns with importance coefficients below $0.01$ to enhance the interpretability of the results.}}
\label{tab:metric_table}
\end{table}

To produce a single number representing the uncertainty for each class, we take the standard deviation of the output $\m Y$ across the MC-Dropout (\textcolor{black}{25 forward iterations or models}) dimension $J$. The generated output $\m S_c$ is a $3-D$ tensor that contains the uncertainty-aware segmentation map for each class for a single test sample as shown in \hyperref[fig:results2]{Fig. 3}. We further sum over the spatial dimensions $H$ and $W$ to obtain a single value for each class. \textcolor{black}{We used the maximum and minimum CNE values to normalize the metric between $0$ and $1$}, as shown in \ref{sec2.3}. The standard deviation will be zero \textcolor{black}{if} there are no discrepancies among predictions of the sampled models, which will generate a small value after summation. Conversely, a non-zero standard deviation will indicate high uncertainty and produce a larger summation. \textcolor{black}{Furthermore, we achieved Expected Calibration Error (ECE) \cite{DBLP:journals/corr/GuoPSW17} value of 0.1317.}

As illustrated in \hyperref[tab:metric_table]{Table 1}, our investigation unveiled that various wetland patterns possess notably high CNE metric values, ranging from 0.8 to 1. These scores signify the existence of top-tier patterns that significantly boost the concept of naturalness. Wetlands are important ecosystems renowned for their roles in carbon storage, safeguarding biodiversity, regulating water resources, and providing niches for unique plant and animal species finely adapted to their specific surroundings.

In contrast, glaciers, grasslands, and water bodies exhibit relatively low-quality patterns, with an approximate metric value of 0.2. These values indicate patterns with a diminished contribution to the naturalness concept, accompanied by heightened uncertainty. 

\section{Conclusion}
We utilize machine learning models to analyze satellite imagery, focusing on understanding naturalness. Our novel approach combines explainable machine learning and uncertainty quantification to provide comprehensive and valid explanations for intricate natural patterns, addressing the limitations of existing methods. Our framework, \textcolor{black}{Confident Naturalness Explanation (CNE)}, offers a quantitative metric and certainty-aware segmentation masks, transforming the understanding of naturalness in Fennoscandia by delivering objective, valid, and quantifiable explanations. In addition, our proposed approach holds promising potential for enhancing protected area preservation and monitoring efforts. \textcolor{black}{ Our results provide a quantifiable index, ranging from 0 to 1, for assessing pattern importance in the context of wilderness. This approach ensures validity by encompassing all distinctive patterns. At the same time, objectivity is maintained through the use of logistic regression coefficients, \textcolor{black}{mitigating hand-engineered features and potential subjectivity and entanglement associated with previous indices and heatmap-based methods.}}

\section*{Acknowledgment}

This work is funded by the Deutsche Forschungsgemeinschaft (DFG, German Research Foundation) RO~4839/5-1 / SCHM~3322/4-1 458156377- MapinWild, RO 4839/6-1 - 459376902 - AID4Crops, and DFG's Excellence Strategy, EXC-2070 - 390732324 - PhenoRob.

\ifCLASSOPTIONcaptionsoff
  \newpage
\fi

\bibliographystyle{ieeetr}
\bibliography{ref}
\end{document}